\documentclass{spie}
\usepackage{graphicx}
\usepackage{verbatim}
\usepackage{subfigure}
\usepackage{setspace}
\usepackage{authblk}
\usepackage{array}
\usepackage{hyperref}
\usepackage{color}
\title{Towards Accurate Camera Geopositioning by Image Matching}
\author{Raffaele Imbriaco, Clint Sebastian,  Egor Bondarev, Peter H.N. de With \\Eindhoven University of Technology, Eindhoven, The Netherlands}
\begin{document}

\maketitle

\begin{abstract}
   In this work, we present a camera geopositioning system based on matching a query image against a database with panoramic images. For matching,  our system uses  memory vectors aggregated from global image descriptors based on convolutional features to facilitate fast searching in the database. To speed up searching, a clustering algorithm is used to  balance geographical positioning and computation time. We refine the obtained position from the query image using a new outlier removal algorithm. The matching of the query image is obtained with a recall@5 larger than 90\% for panorama-to-panorama matching.  We cluster available panoramas from geographically adjacent locations into a single compact representation and observe computational gains of approximately 50\% at the cost of only a small (approximately 3\%) recall loss. Finally, we present a coordinate estimation algorithm that reduces the median geopositioning error by up to 20\%.

\end{abstract}

\section{Introduction}
Radionavigation systems have become increasingly common, with many consumer electronic devices having GPS. However, their positioning accuracy is dependent upon satellite signal quality (affected by multipath, signal blockage, etc). A vision-based geopositioning and navigation system should be unaffected by those factors. Nevertheless, realizing such a system is not trivial, since it requires a large, high-quality collection of pivotal images with accurate positioning information. Furthermore, the system should be capable of rapid and accurate localization under appearance and viewpoint variations. A minimum requirement of such a system is the capability to recognize the location depicted in an image from an arbitrary viewpoint. This problem is known in the computer vision community as visual place recognition.  

Visual place recognition has been traditionally based on hand-crafted local features such as~SURF \cite{bay2008speeded} and SIFT~\cite{lowe1999object}, or on bag-of-words~\cite{sivic2003video}. However, in the past few years there has been a shift towards learned descriptors and features. This is because they leverage the advances in convolutional neural networks (CNNs) to learn descriptors that allow matching of images taken at neighboring locations~\cite{arandjelovic2016netvlad}\cite{babenko2015aggregating}\cite{sunderhauf2015place}, or to directly regress the six degrees of freedom of the camera~\cite{kendall2015posenet}. Methods belonging to the former category are considered to be indirect~\cite{piasco2018survey} and can be treated as an image retrieval task. Given a query image depicting a landmark or a scene, at least one corresponding picture in the database should be found. In our case, the applied scenes are urban landscapes captured with panoramic images. 

Visual place recognition algorithms exploit similarities in feature space to obtain a set of matches for a query image. However, works such as \cite{arandjelovic2016netvlad}\cite{panphattarasap2016visual}\cite{sunderhauf2015place} do not consider additional non-visual information that may facilitate the search process. Physical places are visually consistent, since their appearance does not change abruptly. This means that images acquired at geographically adjacent locations will share a part of their visual features. In a sequence of urban scenes, appearance will change based on the distance between subsequent images, weather conditions and visual clutter (e.g. vehicles). In this work, we adopt the concept of memory vectors to decrease the computational cost and then demonstrate the impact of geographic clustering at different hierarchical levels on the search time and the accuracy. Summarizing, the problem addressed in this paper is to recognize the location of the images based on visual feature vector matching, and then estimate the position of the object of interest as accurately as possible using geographical relations of landmarks in the surrounding scene. 

Our contributions are twofold. First, we present a novel system in which the data areas are hierarchically clustered based on their geographical coordinates to facilitate a high searching speed at city-scale. Second, we present an outlier removal algorithm to remove incorrectly matched image candidates. This algorithm uses geographical and visual similarity relationships between candidate locations to improve the estimated position.

\section{Related work}
In recent years, research regarding image matching has been influenced by the developments in other areas of computer vision. Deep learning architectures have been developed both for image matching \cite{babenko2015aggregating}\cite{iscen2017panorama}\cite{noh2017large}  and geopositioning \cite{lin2015learning}\cite{shan2014accurate}\cite{workman2015wide} with attractive results. 

\textbf{Image matching.}
Convolutional features extracted from the deep layers of CNNs have shown great utility when addressing image matching and retrieval problems. Babenko \emph{et al.}~\cite{babenko2015aggregating} employ pre-trained networks to generate descriptors based on high-level convolutional features used for retrieving images of various landmarks. Sunderhauf \emph{et al.}~\cite{sunderhauf2015place} solve the problem of urban scene recognition, employing salient regions and convolutional features of local objects. This method is extended in~\cite{panphattarasap2016visual}, where additional spatial information is used to increase the algorithm performance. 

\textbf{Geopositioning.}
The problem of geopositioning can be seen as a dedicated branch of image retrieval. In this case, the objective is to compute extrinsic parameters (or coordinates) of a camera capturing the query image, based on the matched georeferenced images from a database. 
There exist many different algorithms and neural network architectures that attempt to identify the geographical location of a street-level query image. Lin \emph{et al.}~\cite{lin2015learning} learn deep representations for matching aerial and ground images. Workman \emph{et al.}~\cite{workman2015wide} use spatial features at multiple scales which are fused with street-level features, to solve the problem of geolocalization. In~\cite{shan2014accurate}, a fully automated processing pipeline matches multi-view stereo (MVS) models to aerial images. This matching algorithm handles the viewpoint variance across aerial and street-level images. 

A common factor of the above work is that it either requires the combination of aerial and street-level images for geopositioning, or extensive training on specific datasets. Both cases and their solutions cannot be easily generalized. In our approach, we utilize georeferenced, street-level panoramic images only and a pre-trained CNN combined with image matching techniques for coordinate estimation. This avoids lengthy training and labeling procedures and assumes street-level data to be available without requiring aerial images. Furthermore, and unlike \cite{iscen2017panorama}, we do not assume that our query and database images originate from the same imaging devices.

\section{Methods}
This section starts with the system architecture of the complete solution. Then, the concept of memory vectors and the proposed processing pipeline (including pre-processing of the panoramic images) are presented. The section also briefly describes the modalities for memory vector aggregation.

\subsection{System architecture.}
\begin{figure}	
\centering           
        \includegraphics[width=0.95\textwidth]{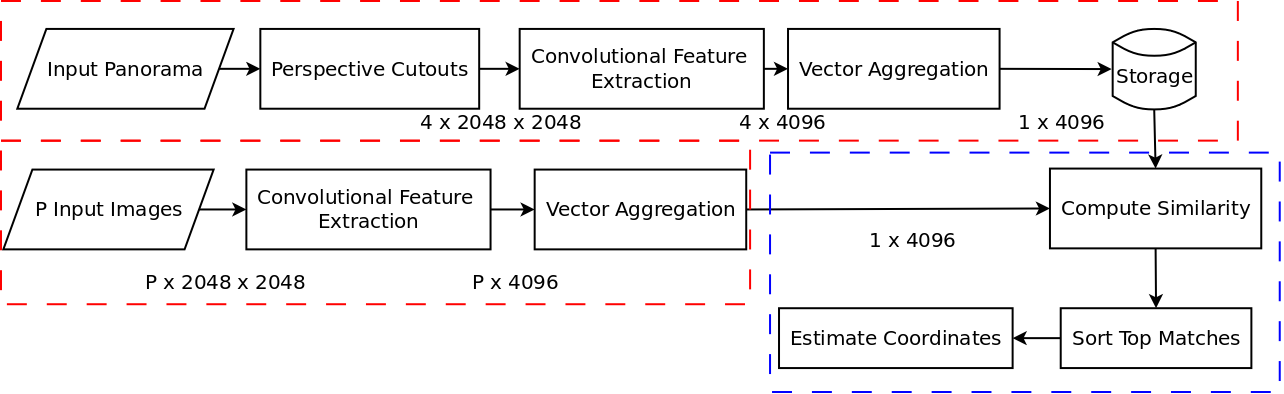}    
    \caption{System diagram. Processes performed offline (for database generation) are within the red box at the top. Processes executed at query time are within the blue box at the bottom.}
    \label{fig:pipe}
\end{figure}

The complete system is depicted in Fig.~\ref{fig:pipe}, which is partitioned into an offline and online processing block. The offline part generates the memory vectors derived from the panoramic images at the input. The description details of these images are given in the experiments section. The bottom part of the architecture refers to online processing and concentrates on querying images. To this end, we extract convolutional features for new input images and generate a feature vector, which is compared to the stored vectors from the offline processing. Some of the processing steps are described below in more detail to specify the involved algorithms and design choices.  

\textbf{Memory vector aggregation.}
Iscen \emph{et al.}~\cite{iscen2017memory}\cite{iscen2017panorama} describe two different ways to aggregate feature vectors into memory vectors. The first one consists of summing the feature vectors. The second method uses a weighted sum for the aggregation of the vectors. The weights are obtained from the Moore-Penrose pseudo-inverse. We adopt the naming convention of \cite{iscen2017panorama} and refer to the sum memory vector as \textit{sum vector} and to the pseudo-inverse memory vector as \textit{p-inv vector}. Once each panoramic picture has been processed, a database is generated containing the corresponding memory vectors. 

\textbf{Feature extraction and matching.}
In \cite{iscen2017panorama} the query and database images are acquired by the same device. This means that while changes in appearance and viewpoint are included in the query images, additional variations introduced by different camera sensors are occurring but not considered. Furthermore, in image retrieval tasks, it is common to consider the database as a collection of isolated images. However, geotagged image collections of urban scenes (such as our dataset or Pittsburgh250K~\cite{torii2013visual}) contain additional information that can be used for efficient searches in city-scale databases. We perform aggregation both in feature and geographical space. Similarly to~\cite{iscen2017panorama}, we use NetVLAD~\cite{arandjelovic2016netvlad} as the feature extractor. Due to the size of our panoramic images, direct feature vector extraction is computationally too expensive. Our approach therefore decomposes the spherical images into four planar images, corresponding to each cardinal direction.

Matches are found by calculating the cosine similarity between the query memory vector and each stored memory vector in the database. Then, the candidates with the highest similarity scores are chosen. To reduce the number of searches at query time, while exploiting the fact that the data from our main source is acquired every 5~meters, we cluster the images based on their geographical coordinates. For this, a cluster of panoramic images is  grouped into regularly-sized neighborhoods, using agglomerative clustering~\cite{rokach2009survey} and afterwards, a single memory vector summarizing the cluster is generated. In this way, the number of searches can be reduced by a factor of $N$$\times$$M$ where $N$ denotes the cluster size and $M$ is the number of times the clustering is performed (hereafter referred to as granularity). Thus, granularity zero represents a full-search through all images and granularity $M$ requires clustering and aggregating the data $M$ times in a hierarchical structure. If the granularity is larger than zero, the search process will be repeated within each candidate cluster until the candidates are panoramic images and we have arrived at granularity zero. 

\textbf{Coordinate estimation.}
The set of matching images at the output of the system may contain false positives that do not correspond to the query location. A high similarity score does not necessarily mean that the locations are exactly the same. They could share a significant amount of features (surrounding vegetation, large objects of similar appearance), and yet sometimes be even geographically separated. Thus, a naive approach that calculates an average or the center of gravity from the coordinates of the matches would result in inaccurate positioning. An additional contribution in our approach is that we re-rank the candidates based on their cross-similarity and their relative geographical distance to remove erroneous matches.

After matching, a set of panoramic pictures $\mathbf{P}$ = $\{P_0, P_1, ..., P_N\}$, their corresponding coordinates $C$ and similarity scores $S$ are obtained. For an arbitrary picture $P_i$ that is matched, the similarity score to each other picture $P_j$ of $\mathbf{P}$ is calculated and divided by their geographical distance. Our coordinate-estimation algorithm is as follows. The similarity scores with the individual other pictures are accumulated and defined as the new rank $r_i$ of the picture. The mean of the new ranks of the images matching the set $\mathbf{P}$ is calculated and is used as a threshold. The individual elements of $P_j$ with a rank below the mean are rejected. Finally, we calculate the center of gravity of the remaining pictures of the set $\mathbf{P}$ after one matching cycle, using the original similarity score as the mass. Ranks $r_i$ are specified by: 
\begin{equation}
r_i = \sum_{i \neq j}^{n} \frac{\mathbf{S}(P_i, P_j)}{\|c_i - c_j\|_2}, 
\label{rerank_equation}
\end{equation}
where $\|c_i - c_j\|$ denotes the Euclidean distance between the geographical locations.

\section{Setup of Experiments}
This section addresses the experimental details of our geopositioning algorithm. We present our datasets, the metrics employed and the image matching and geopositioning experiments performed.

\textbf{Data and evaluation criteria.}
The acquisition of high-resolution panoramic images of selected European and American cities is performed by several commercial companies. For our experiments, a Dutch company\footnote{Cyclomedia is a company that sells annually recorded pictures to government and civil engineering agencies.} has provided UHD $360^{\circ}$ spherical panoramas, covering a large part of the city of Eindhoven and their corresponding coordinates. The dimensions of these images are 14k$\times$7k pixels. Alternative to this, street-level images from the Internet\footnote{These images are extracted from the Google Streetview database.} are also available. 

Our dataset consists of 39,333 georeferenced panoramic images (abbreviated further as C-images). Per image, four planar subimages of 2048$\times$2048~pixels are generated, corresponding to the four cardinal orientations. We uniformly select 500 panoramic images to use as the test set excluding them from the database. We also test on images generated outside the dataset, but depicting the same location. To this end, we have also obtained 250 panoramic images from street-level source (further abbreviated as G-images) at randomly selected locations matching to those in our dataset. There are significant differences between the panoramic images obtained by both sources. The contents of the locations (people, vehicles, signs, etc.) differ, as does the weather conditions and the exact position of the acquisition vehicles on the road. G-images are treated equally as C-images, but are used exclusively as queries.

\textbf{Finding images with the best similarity score.}
In our experiments, we attempt to identify panoramic images that visualize the closest location to the query image. For the database, we only generate memory vectors using all four subimages of the entire panoramic image. We evaluate the accuracy of the matches when only one planar image and all planar images are used as a query. We abbreviate these cases as Im2Pan and Pan2Pan, respectively. For this, we compute the similarity score per query against the whole database using the cosine distance. We repeat this procedure with feature vectors commonly used in image retrieval tasks and compare the results. These are VGG16~\cite{simonyan2014very}, R-MAC and MAC \cite{tolias2015particular}. We report the mean recall@N, the same metric used for the Pittsburgh250K dataset, where a matching set is considered correct if it contains at least one image found within 25~meters of the query location. These tests are also repeated at different granularity levels of geographical clustering as discussed earlier. We measure the impact of the cluster size on the overall matching accuracy and computation time. For these experiments, we study three granularity levels (GL0, GL1 and GL2) with cluster sizes of 4, 8 and 16 panoramic images.
 
 \textbf{Matching images outside the dataset.}
The previous experiments assume that both the query and database images are taken by identical systems under similar conditions. However, in real life, this is rarely the case. We compare the retrieval performance when the query images are acquired with different sensors and under different conditions, by using C-images as the database and G-images as the queries. The objective of these experiments is to demonstrate the capabilities of our system to generalize and simulate a more realistic application.
 
\textbf{Coordinate estimation.}
Additionally, we calculate the error between the estimated and ground-truth coordinates of the original query. This is computed as the Euclidean distance of the coordinates in the Dutch (``Rijksdriehoek") coordinate system. For these experiments, we first perform image matching for all the different subimages. From this set, a georeferenced position is generated by computing the center-of-mass estimation. We also generate coordinate estimates after removing outliers with our method and compare those against the estimate obtained from the original matching set.

\section{Results and discussion}
The following results are only based on using the \textit{p-inv vector}, as it continuously scores better than the \textit{sum vector}. This is in accordance with experiments from literature. In this section we conduct the following experiments. First, we employ either one planar image (Im2Pan) or four aggregated planar images as a query (Pan2Pan). In both cases, we use the same database with C-images and concentrate on the quality of the matching process for evaluation. This quality is expressed as the mean recall@N. Then in the second experiment, we simulate a real application using G-images as query input. Finally, in the third experiment we study the accuracy of our geopositioning estimates. 

\begin{table}
	\centering
  	\caption{Average recall@N for various convolutional descriptors. Query and database images are acquired with the same device. Matching is done with multi-core processing on an Xeon CPU at 2.20G~Hz.}		
    \label{tab:pan2pan} 
\begin{tabular}{|l|c|c|c|c|}
\hline 
 & \multicolumn{3}{c|}{\textbf{Avg. recall@N (in \%) }} & \textbf{Avg. time (in ms)} \\ \hline
  & \textbf{Top 1} & \textbf{Top 5} & \textbf{Top 10} & \\ \hline 
Im2Pan - VGG16 & 44.97 & 72.56 & 82.16 & 1.59\\
Im2Pan - MAC & 37.70 & 57.50 & 65.35 & 0.45\\
Im2Pan - R-MAC & 54.30 & 77.00 & 83.55 & 0.44 \\
Im2Pan - NetVLAD & 68.13 & 85.20 & 90.25 & 1.80\\ \hline
Pan2Pan - VGG16 & 63.62 & 97.56 & 98.17 & 1.69\\
Pan2Pan - MAC & 82.40 & 92.20 & 96.60 & 0.45\\
Pan2Pan - R-MAC & 96.60 & 99.20 & 99.60 & 0.44\\
Pan2Pan - NetVLAD & 95.10 & 100 & 100 & 1.61\\ \hline
\end{tabular}
\end{table}

\begin{figure}[h]
	\centering
  	\caption{Average recall@N for the best performing descriptors after geographical clustering. Two granularity levels and three cluster sizes are shown.}		
    \label{fig:clustering} 
\includegraphics[width=\linewidth]{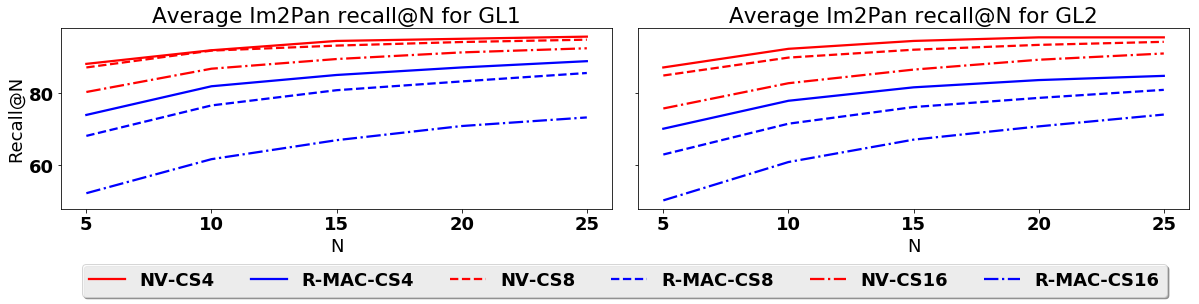}
\end{figure}

\textbf{Exp. 1: Image matching from similar sources.}
Table \ref{tab:pan2pan} summarizes the results of the Im2Pan and Pan2Pan experiments. In the simpler Pan2Pan case, we notice similar performance across the different descriptors. However, we also observe that NetVLAD significantly outperforms the other three convolutional descriptors in the harder Im2Pan case. The performance difference observed is due to the specialized design and training of NetVLAD. The other convolutional descriptors are extracted from CNNs trained for classification. Meanwhile, NetVLAD has been specifically trained for place recognition. This means that NetVLAD has learned features that are better suited for describing urban scenes, while the other descriptors exhibit more generic features. With regards to geographical clustering, we observe a trade-off between the recall and computation time. Each hierarchical clustering step halves the computation time. However, this improvement is at the expense of the matching accuracy. The loss in recall is dependent on the granularity level and number of images per cluster. As the clusters become larger, the recall drops accordingly. This behaviour is depicted in Figure~\ref{fig:clustering}, where a larger $N$ for clustering moves the recall curve downwards. 

\textbf{Exp 2: Image matching from different sources.}
This experiment simulates a real system where the query and the database images originate from different sources. We report a performance drop of over 30\% for all descriptors, as seen in Table~\ref{tab:gimage}. The most likely cause of this performance degradation is the quality difference between the query and database images. G-images capturing the same location may have a significantly different visual appearance due to several factors. Among these, we list here the weather conditions, large view-point differences, scene clutter (vehicles, pedestrians, etc.) and imaging artifacts. Each of these factors has an impact on the memory vector generated from the image, which complicates the matching task. 

\begin{table}[h]
	\centering
  	\caption{Average recall@N for various convolutional descriptors. Query and database images are acquired with different devices under different conditions.}		
    \label{tab:gimage} 
\begin{tabular}{|l|c|c|c|c|c|}
\hline 
 & \multicolumn{5}{c|}{\textbf{Average recall@N (in \%) }}  \\ \hline
  & \textbf{Top 1} & \textbf{Top 5} & \textbf{Top 10} & \textbf{Top 15} & \textbf{Top 20} \\ \hline 
Im2Pan - VGG16 & 0.00 & 0.20 & 0.70 & 1 & 1.3\\
Im2Pan - MAC & 7.00 & 15.20 & 19.55 & 23.30 & 25.45\\
Im2Pan - R-MAC & 2.25 & 9.3 & 18.05 & 24.30 & 29.35 \\
Im2Pan - NetVLAD & 30.85 & 53.00 & 60.05 & 64.05 & 66.70\\ \hline
\end{tabular}
\end{table}

\begin{table}
	\centering
    \caption{Geopositioning results expressed in obtained distance from the real query location (in meters) for various convolutional descriptors for Im2Pan and Pan2Pan and  matching-set sizes.}
    \label{tab:geo_error}
	\begin{tabular}{|c|c|c|c|c|c|c|c|c|}
	\hline
     & \multicolumn{4}{c|}{\textbf{Baseline - Im2Pan}} & \multicolumn{4}{c|}{\textbf{Re-ranked - Im2Pan}}\\ \hline
     & \textbf{Top 5} & \textbf{Top 10} & \textbf{Top 15} & \textbf{Top 20} & \textbf{Top 5} & \textbf{Top 10} & \textbf{Top 15} & \textbf{Top 20} \\ \hline
    MAC & 601.15 & 696.94 & 707.68 & 710.23 & 489.69 & 501.67 & 568.20 & 568.75\\ \hline 
    R-MAC & 307.01 & 390.96 & 498.82 & 523.02 & 293.74 & 378.74 & 447.63 & 488.04 \\ \hline
    NetVLAD & 103.59 & 238.36 & 304.91 & 335.12 & 97.98 & 182.95 & 248.82 & 307.51\\ \hline
    & \multicolumn{4}{c|}{\textbf{Baseline - Pan2Pan}} & \multicolumn{4}{c|}{\textbf{Re-ranked - Pan2Pan}}\\ \hline
     & \textbf{Top 5} & \textbf{Top 10} & \textbf{Top 15} & \textbf{Top 20} & \textbf{Top 5} & \textbf{Top 10} & \textbf{Top 15} & \textbf{Top 20} \\ \hline
         MAC & 75.64 & 227.97 & 296.27 & 326.68 & 60.04 & 165.74 & 205.45 & 242.56\\ \hline 
    R-MAC & 10.43 & 51.17 & 117.84 & 158.58 & 10.78 & 41.05 & 77.28 & 113.98 \\ \hline
    NetVLAD & 8.88 & 28.77 & 56.15 & 93.59 & 8.96 & 27.41 & 47.11 & 75.61\\ \hline
	\end{tabular}
\end{table}
\pagebreak

\textbf{Exp. 3: Geopositioning.}
The geopositioning results are presented in Table~\ref{tab:geo_error} and use C-image data as both query and database. This experiment is constrained to C-images, due to the poor performance in the previous experiment. When we use our re-ranking approach, we notice a reduction in the median geopositioning error, when compared against the original matching set. For the Pan2Pan and Im2Pan cases, we achieve an error reduction of approximately 20\% and  6\%, respectively. Nevertheless, our approach has limitations, since it  depends on the accuracy of the matching set. Difficult cases include complete mismatches or candidate sets with an incorrect and yet geographically adjacent majority.

\section{Conclusion}
We have presented an image-based geopositioning system based on image matching techniques. Our system uses global image descriptors based on convolutional features stored in memory vectors and an effective re-ranking algorithm to estimate the geographic coordinates of an image. Experimental results demonstrate the computational cost reduction of geographical clustering. We achieve a reduction of roughly 50\% in computation time at a minimal drop in recall (approximately 3\%). This clearly proves the feasibility of our clustering algorithm. We accurately match to panoramic images of the locations, obtaining a recall@10 larger than 90\% for panorama-to-panorama matching. We have also reported on the difficulty of matching a planar image to a complete panoramic image using memory vectors and explored generalized image matching. We conclude that visually similar images may still produce different memory vectors, which hampers correct location recognition. This becomes evident when the query image is acquired under different conditions by different devices. A likely cause of this behavior is visual clutter (e.g. cars, pedestrians, vegetation). Further research is required for reducing the impact of visual clutter on location recognition.  

Regarding geopositioning, we demonstrate that our method reduces the median geopositioning error by up to 20\% using our outlier removal algorithm. We also show that its accuracy is affected by the results of the underlying image matching system. Our approach can produce estimates within 20~meters of the original query location, depending on the number of candidate matches and number of query planar images. The quality of the estimation also depends on the accuracy of the matches, since incorrect majorities or a large geographical spread among the matches, produces inaccurate estimates. The latter issue can be addressed by geometric verification of local features. 

\pagebreak
\bibliography{main_final}
\bibliographystyle{abbrv}

\pagebreak
\section*{Author's Background}
\begin{table}[h]
\centering
\begin{tabular}{|c|c|c|c|}
\hline 
	Your Name & Title & Research Field & Personal Website \\ \hline
    Raffaele Imbriaco & Ph.D. Candidate & Deep Learning, Visual Place recognition & {\color{blue}\href{http://vca.ele.tue.nl/user/102}{Website}} \\ \hline
    Clint Sebastian & Ph.D. Candidate & Deep learning on 3D point clouds & {\color{blue}\href{http://vca.ele.tue.nl/user/113}{Website}} \\ \hline
    
    Egor Bondarev & Assistant Professor & 3D Reconstruction, Video Content Analysis & {\color{blue}\href{http://vca.ele.tue.nl/user/28}{Website}} \\ \hline
    Peter H.N. de With & Full Professor & Video Coding, Analysis and Architectures & {\color{blue}\href{http://vca.ele.tue.nl/user/18}{Website}} \\ \hline
    
\end{tabular}
\end{table}

\end{document}